\documentclass[runningheads]{llncs}
\usepackage{graphicx}
\usepackage{dirtytalk}
\usepackage{float}
\usepackage{multirow}
\usepackage[table,xcdraw]{xcolor}
\usepackage{amssymb}
\usepackage{tabularx}
\usepackage{bbding}
\usepackage{cite}
\usepackage[colorlinks=true]{hyperref}
\hypersetup{
    linkcolor=red, 
    urlcolor=green,
}

\begin{document}
\title{Design as Desired: Utilizing Visual Question Answering for Multimodal Pre-training}

%
\titlerunning{Design as Desired: Utilizing VQA for Multimodal Pre-training}
%
\author{Tongkun Su\inst{1,2,*} \and
Jun Li\inst{3,}\thanks{The first two authors contributed equally to this work.} \and
Xi Zhang\inst{1,2} \and Haibo Jin\inst{5} \and Hao Chen\inst{5} \and \\ 
Qiong Wang\inst{1,6} \and Faqin Lv\inst{4} \and Baoliang Zhao\inst{1(}\Envelope\inst{)} \and Yin Hu\inst{1(}\Envelope\inst{)}} 
\institute{Shenzhen Institute of Advanced Technology, Chinese Academy of Science
\and
University of Chinese Academy of Science
\and
Technical University of Munich, Munich Center for Machine Learning 
\and
Southern Medical University
\and
The Hong Kong University of Science and Technology
\and
The Chinese University of Hong Kong
}
\maketitle             

\begin{abstract}

Multimodal pre-training demonstrates its potential in the medical domain, which learns medical visual representations from paired medical reports. However, many pre-training tasks require extra annotations from clinicians, and most of them fail to explicitly guide the model to learn the desired features of different pathologies. To the best of our knowledge, we are the first to utilize Visual Question Answering (VQA) for multimodal pre-training to guide the framework focusing on targeted pathological features. In this work, we leverage descriptions in medical reports to design multi-granular question-answer pairs associated with different diseases, which assist the framework in pre-training without requiring extra annotations from experts. We also propose a novel pre-training framework with a quasi-textual feature transformer, a module designed to transform visual features into a quasi-textual space closer to the textual domain via a contrastive learning strategy. This narrows the vision-language gap and facilitates modality alignment. Our framework is applied to four downstream tasks: report generation, classification, segmentation, and detection across five datasets. Extensive experiments demonstrate the superiority of our framework compared to other state-of-the-art methods. Our code is available at \url{https://github.com/MoramiSu/QFT-MICCAI2024}.


\begin{figure}[h!]
    \centering
    \includegraphics[width=0.9\textwidth]{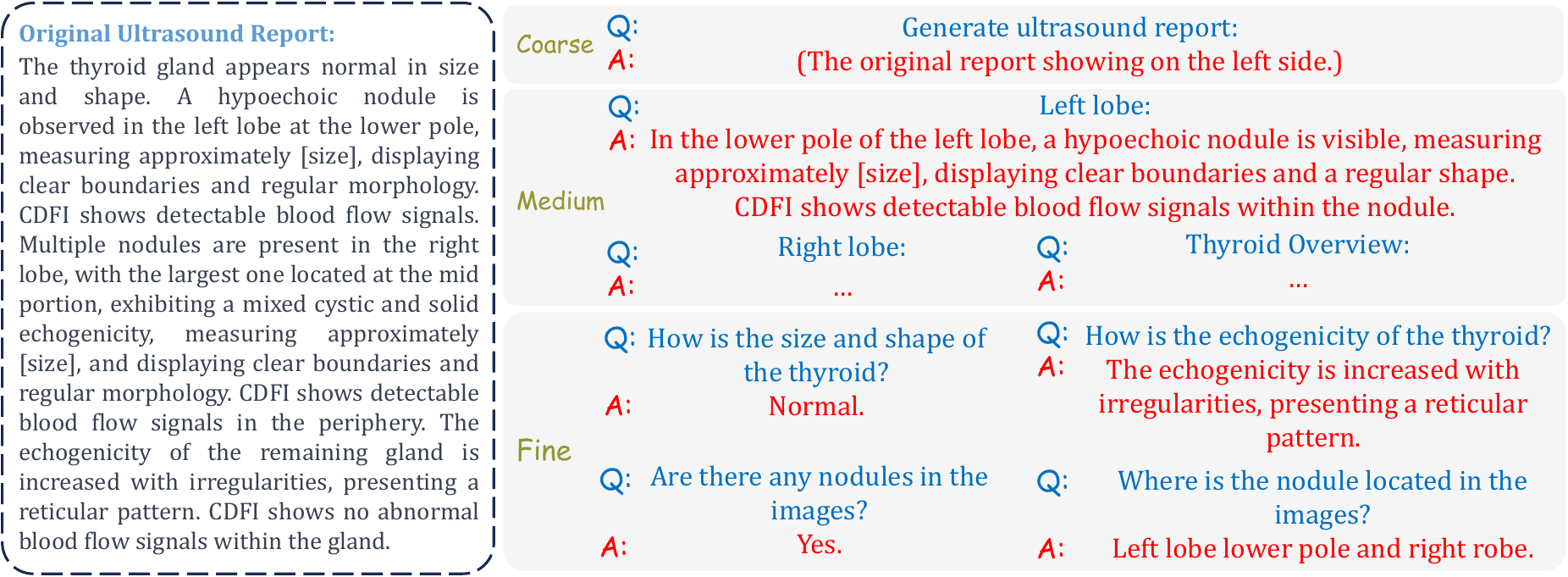}
    \caption{Overview of the VQA design. Left: the original report. Right: question and answer pairs. We design three levels of VQA to enable the model to capture multi-granular features. Numerical values are replaced with a special token [size].}
    \label{VQA}
\end{figure}
\end{abstract}

\section{Introduction}

In recent years, multimodal pre-training has received significant attention. In clinical settings, there is a vast amount of medical images and text reports stored in the database, making them easily available for pre-training tasks. Multimodal pre-training aims to learn generalized representations through the inherent relationships between image-report pairs to benefit various downstream tasks, particularly where annotated data is limited. Generally, there are two mainstream methods in multimodal pre-training: contrastive learning  \cite{MedCLIP,MGCA,GLoRIA,CMITM} and cross-modal reconstruction \cite{MedIM,MRM,CMITM}. The first method aims to learn discriminative representations by pulling closer the representations from paired images and reports while pushing away the unpaired ones. The second method generates one modality from the other, which assists the framework in understanding the relationship between two modalities. However, these methods have some drawbacks. On the one hand, some methods \cite{MedCLIP,MedIM} require extra annotations from clinicians, which is often time-consuming and cost-prohibitive. On the other hand, unlike supervised learning, where labels guide the model to learn the features of interest, self-supervised methods do not explicitly direct the model to focus on the specific features associated with different diseases.

Therefore, our objective is to explore an effective medical multimodal pre-training task that can guide the model to focus on the desired features of different pathologies without the need for further annotations by clinicians. Our attention turns to VQA, an essential task in cross-modal generation which requires the model to understand both visual and textual knowledge\cite{CAT-ViL,WSDAN,MedVQA,MedVQALLM}. Clinicians just need to provide the pathologies they care about in the reports, and we can design different questions according to their instructions. By answering the questions during pre-training, the framework will attempt to focus on different levels of information according to the questions. Figure \ref{VQA} illustrates the main idea of question-answer design. We design VQA tasks at three levels of granularity to guide the model in learning essential details in the image and text.

Additionally, we propose a Quasi-textual Feature Transformer (QFT) module with a contrastive learning strategy to help the framework bridge the gap between image and text. This challenge stems from the fact that discriminative pathological features and tokens often occupy only a small fraction of the medical images and reports, making it difficult for the model to learn their associations effectively. Inspired by QFormer \cite{BLIP2}, we propose a QFT module, which utilizes contrastive learning to convert visual features into a quasi-textual domain that is closer to the textual domain. This transformation narrows the distribution gap between two modalities and improves the model's visual understanding capabilities. Moreover, we construct an ultrasound dataset with different organs for pre-training, comprising 10,720 ultrasound images and 5,360 reports. We transfer our model to four downstream tasks: report generation, classification, detection and segmentation. Extensive experiments demonstrate the superior performance of our framework compared to other State-Of-The-Art (SOTA) methods. Overall, our main contributions can be summarized as:  
\begin{itemize}
    \item To the best of our knowledge, we are the first to utilize VQA for multimodal pre-training in the medical field to assist the framework in focusing on the desired pathological features without extra expert annotationss.
    \item We propose a  QFT module with a contrastive learning strategy, which aligns the visual features into a quasi-textual domain to narrow the modality gap and facilitate modality alignment.
    \item Our approach demonstrates significant improvement in four downstream tasks: report generation, classification, detection and segmentation.
\end{itemize}

\section{Methods}
\label{methods}

\begin{figure}[t!]
    \centering
    \label{QFT}
    \includegraphics[width=\textwidth]{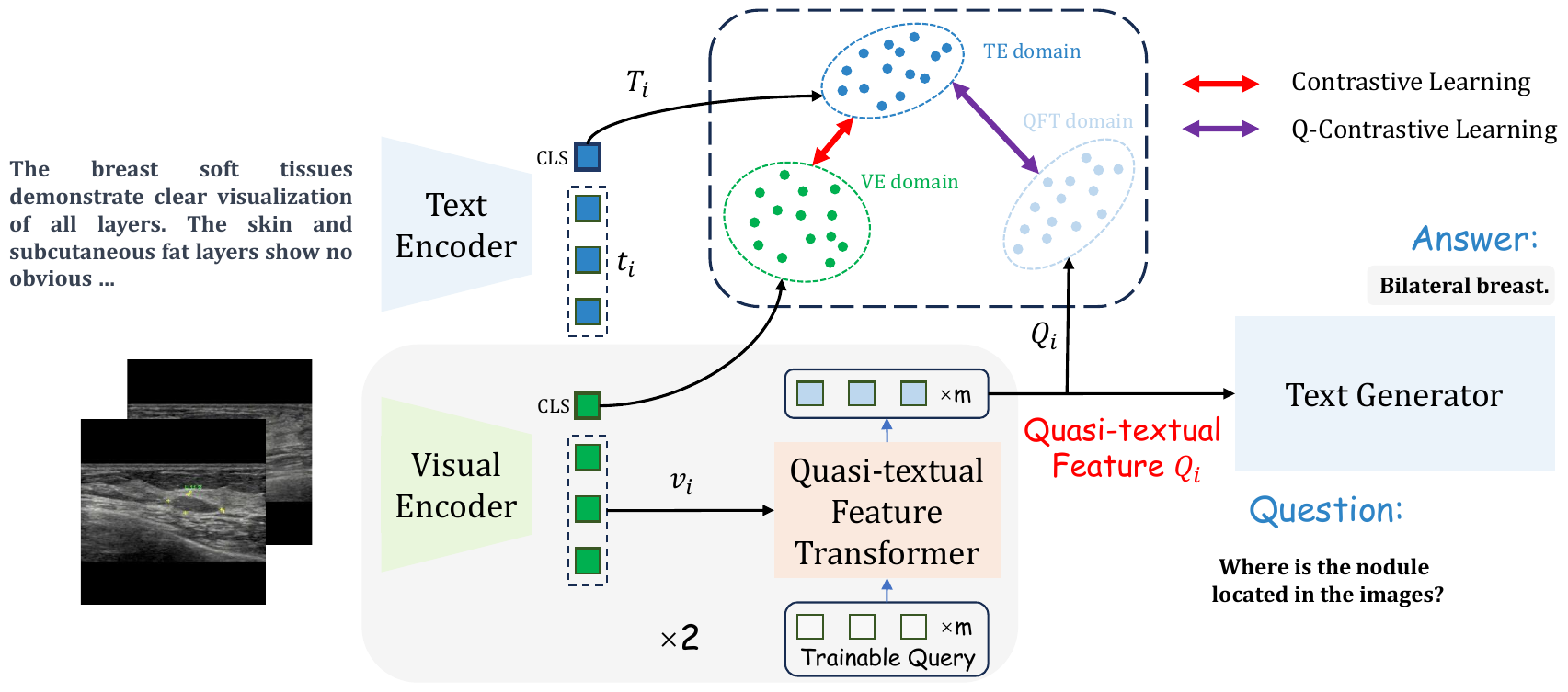}
    \caption{Overview of our framework. The image and text features extracted by the visual and text encoders are aligned by the quasi-textual Feature transformer module with two contrastive learning tasks. Then the quasi-textual features are fed to the text generator to generate answers based on the questions during pre-training. TE domain, VE domain, QFT domain denote the latent space of global textual features $T_i$, global visual features $V_i$ and quasi-textual features $Q_i$, respectively.}
\end{figure}

\subsubsection{Overview.}
We aim to conduct multimodal pre-training through VQA, enabling the model to learn multi-granular pathological information, and thereby enhancing the performance across various downstream tasks. We employ an encoder-decoder architecture as the backbone of our model. Given a mini-batch of images and their corresponding question-answer pairs with the batch size of $B$, we encode the images using the visual encoder (\textit{e.g.}, ViT \cite{ViT} or ResNet \cite{ResNet}) to obtain global features $V_i$ and patch features $v_i, i\in B$. Subsequently, the text generator (\textit{e.g.}, GPT \cite{GPT}) generates answers based on the obtained image features and questions. Inspired by QFormer \cite{BLIP2}, we propose a Quasi-textual Feature Transformer (QFT) module on top of the general encoder-decoder framework to facilitate modality alignment. Our model also includes a text encoder (\textit{e.g.}, BERT \cite{BERT}), which encodes reports into global features $T_i$ and token features $t_i$. Our framework is illustrated in Figure \ref{QFT}.

\subsubsection{Quasi-Textual Feature Transformer.}
\label{QFT section}
In the medical field, discriminative pathological features generally occupy only a small fraction of the medical images, which poses a challenge for modality alignment. To narrow the gap between image and text, we propose a Quasi-textual Feature Transformer (QFT) module with a contrastive learning strategy. The backbone of QFT is a multi-layer bidirectional transformer decoder \cite{transformer}. It takes $m$ trainable queries as input. The queries interact with patch features $v_i$ through the cross-attention mechanism. To enforce the queries to extract quasi-textual features $Q_i$ that are closer to the textual domain, we implement a Q-Contrastive Learning (QCL) for the QFT module.
We calculate the pairwise similarity between global textual features $T_i$ and all $m$ tokens in the output $Q_i$ from the QFT module, selecting the highest score in the similarity matrix as the final score, which can be formulated as $s_q(Q_i, T_i) = \max_{0\leq l \leq m} s(Q_i[l], T_i)$, where $s(\cdot, \cdot)$ computes the cosine similarity, $Q_i[l]$ is one of the output token in $Q_i$. Then we use this similarity score to calculate the InfoNCE loss \cite{InfoNCE}. By minimizing this loss, the model will align the latent space of $Q_i$ with that of $T_i$ together. Given that medical images of the same organ share similar features \cite{SGF}, leading to significant visual redundancy, we introduce a bottleneck design into our model. we set the $m$ much smaller than the number of visual tokens $v_i$. This bottleneck requires the QFT module to compress visual information and extract the most robust information, while also reducing the computational cost of the text generator.




Additionally, we introduce the vision-language Contrastive Learning (CL) to enhance the model's visual perception. As mentioned in \cite{MME}, many QFormer-based models perform poorly on visual perception tasks, especially in position recognition, which is vital in the medical domain. We assume that this phenomenon is because these models all apply a pre-training process similar to QCL. Although QCL facilitates alignment between different modalities, directly pulling $Q_i$ to the textual domain will lead to a significant loss of fine-grained visual information. Thus we use contrastive learning as a constraint to further align the latent space of visual features $V_i$ and that of textual features $T_i$ together, ultimately maintaining more visual features in $Q_i$. Moreover, following \cite{ALBEF}, we introduce three buffers with a buffer size of $N$ for QCL and CL. We store the global features $T_i$, $V_i$ and quasi-textual features $Q_i$ from the most recent batches as negative samples for the current batch. Thus, each batch has $N+B$ negative samples. The QCL and CL loss can be formulated as below, where $\tau_q$ and $\tau_c$ are the temperature parameter.

{\small
\begin{equation}
     \mathcal L_{qcl} = -\frac{1}{2B}\sum_i^B\left[log\frac{\exp(s_q(Q_i, T_i)/\tau_q)}{\sum_{j=1}^{N+B} \exp(s_q(Q_i, T_j)/\tau_q)} + log\frac{\exp(s_q(Q_i, T_i)/\tau_q)}{\sum_{j=1}^{N+B} \exp(s_q(Q_j, T_i)/\tau_q)}\right]
\end{equation}
\begin{equation}
    \mathcal L_{cl} = -\frac{1}{2B}\sum_i^B\left[log\frac{\exp(s(V_i, T_i)/\tau_c)}{\sum_{j=1}^{N+B} \exp(s(V_i, T_j)/\tau_c)} + log\frac{\exp(s(V_i, T_i)/\tau_c)}{\sum_{j=1}^{N+B} \exp(s(V_j, T_i)/\tau_c)}\right]
\end{equation}}

\subsubsection{Vision Question Answering.}
According to the data we gathered from the hospital, each report is associated with two images. Thus, we encode two images separately with a shared visual encoder and the QFT module. We take the average of two global visual features $V_i$ for CL, the average of two quasi-textual features $Q_i$ for QCL, and the concatenation of $Q_i$ for generation. For each pair of medical images, we randomly sample three question-answer pairs from coarse, medium, and fine-grained VQA tasks respectively. We then use the questions as prompts to guide the generation. Given the question $p_i$ and ground truth answer $y_i = \{y_{i, 1}, y_{i, 2}, ... y_{i, L}\}$ composed of $L$ tokens, the language modeling loss can be formulated as:
\begin{equation}
    l_{clm, mlm, flm} = \frac{1}{BL}\sum_i^B\sum_j^L \left[-log(p(y_{i,j}|y_{i,<j}, p_i, q_i)) \right]
\end{equation}
\begin{equation}
    \mathcal L_{lm} = \lambda_c l_{clm} + \lambda_m l_{mlm} + \lambda_f l_{flm}
\end{equation}
where $l_{clm, mlm, flm}$ represent language modeling loss of coarse, medium and fine-grained VQA. $\lambda_c$, $\lambda_m$, $\lambda_f$ are hyperparameters. We apply ViT-B to initialize the visual encoder, the first six layers of ERNIE\cite{ERNIE-health-zh} to initialize text encoder, and GPT2-small \cite{GPT2,UER,Tencentpretrain} to initialize the text generator. For the QFT module, we initialize it with the first three layers of GPT2-small to equip it with the same knowledge as the text generator. We weight CL with $\lambda$ to balance the constraint. Our loss function is shown as below:
\begin{equation}
    \mathcal L = \lambda \mathcal L_{cl} + \mathcal L_{qcl} + \mathcal L_{lm}
\end{equation}

\section{Experiments}
\subsubsection{Dataset.}
We construct a dataset for experiments with 10,720 ultrasound images and 5,360 Chinese reports for breast and thyroid. The dataset is divided into training, validation, and test sets in a ratio of 7:1:2. The training and validation sets are used for pre-training, while the test set is used for downstream report generation. For three visual recognition tasks, we evaluate the effectiveness of pre-training on three public datasets: BUSI \cite{BUSI}, AUITD \cite{AUITD}, and DDTI \cite{DDTI}. 
\subsubsection{VQA Design.}
Figure \ref{VQA} shows the overview of our VQA design (the texts are translated from Chinese to English by ChatGPT \cite{ChatGPT}). In coarse-grained VQA, we require the model to generate complete reports. During this process, the model will focus on the report formats and writing styles. However, the medical reports are usually very long and templated, so coarse-grained VQA is not enough to highlight the details in the reports. Therefore, we further design medium and fine-grained questions. In medium-grained VQA, we require the model to generate clinical descriptions for various anatomical regions, enabling the model to effectively differentiate anatomical structures. In fine-grained VQA, we design a series of questions based on the report, targeting subtle but crucial pathological visual features.  Given that nodule recognition is the most important in ultrasound, our designed questions are mainly focusing on the presence and location of the nodules. Finally, we obtain a VQA dataset with 23,572 question-answer pairs. More examples are presented in the supplementary material.

\subsubsection{Framework Setup.}
Considering that report generation is a generative task, while visual recognition is a discriminative task, we separately train the framework with two settings for different downstream tasks. For the report generation, since we utilize it as coarse-grained VQA in pre-training, we directly utilize the prompt in coarse-grained to require the model to generate reports. We set $\lambda_{c} = 9$, $\lambda_{m} = 1$, and $\lambda_f = 3$ to emphasize coarse-grained VQA. We use BLEU scores \cite{BLEU}, METEOR \cite{METEOR} and ROUGE-L \cite{ROUGE-L} to evaluate report quality. For the other three visual recognition tasks, to guide the model focusing on the pathological features emphasized in the middle and fine-grained VQA, we set $\lambda_{c} = 1$, $\lambda_{m} = 3$, and $\lambda_f = 9$. We transfer the pre-trained visual encoder into different downstream backbones (\textit{e.g.}, YOLOv3 \cite{YOLOv3} for detection) and only fine-tune the other parts of the backbone. We compare accuracy (ACC) for classification, average precision (AP) for detection and DICE score for segmentation. We set $\lambda = 1$ in both settings. All baselines are retrained on our ultrasound dataset. More framework setup details are illustrated in supplementary materials. 

\section{Results and Analysis}

\begin{figure}[t!]
    \centering
    \includegraphics[scale=0.31]{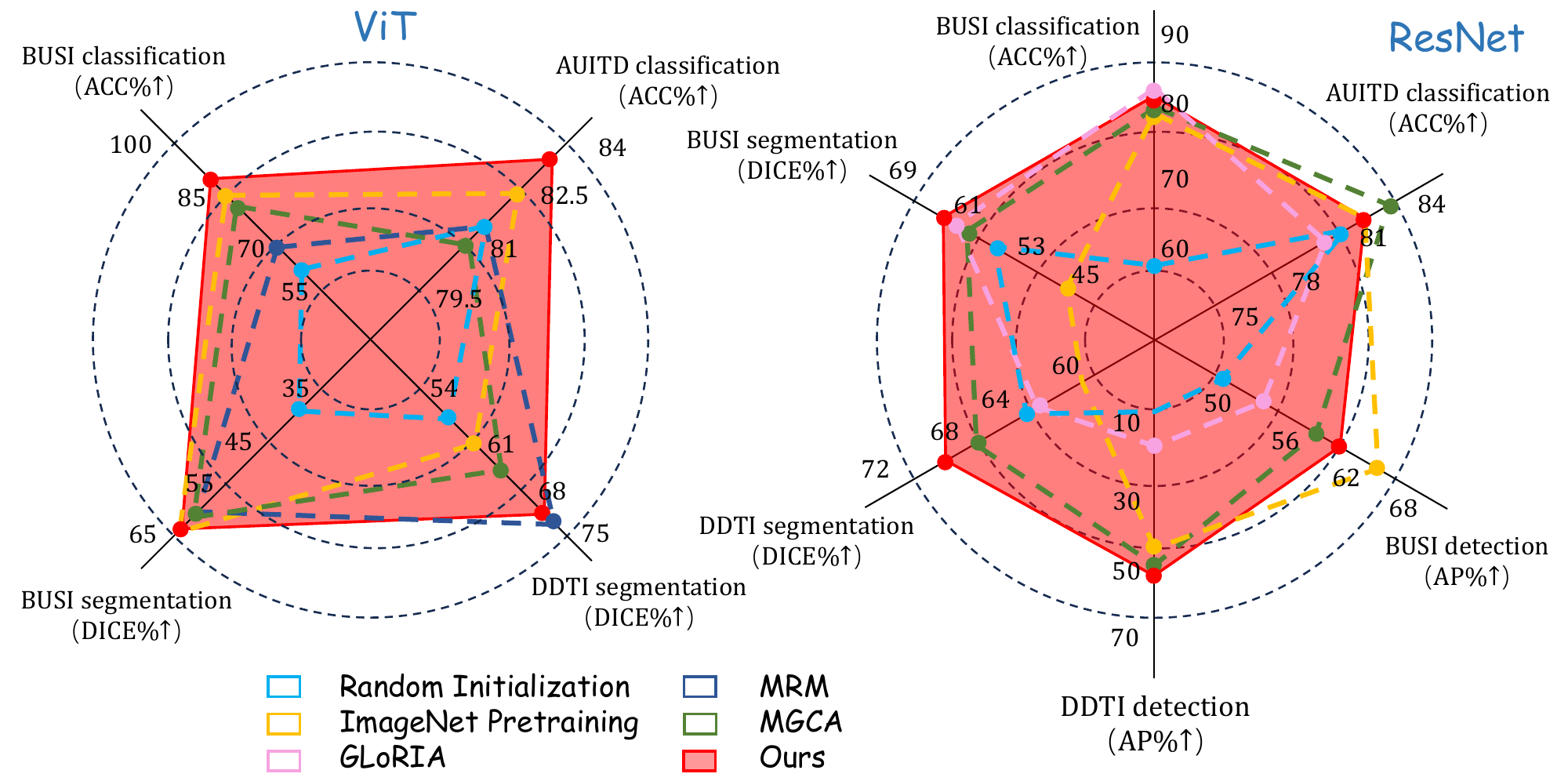}
    \caption{Performance on visual recognition tasks. We compare our method with GLoRIA \cite{GLoRIA}, MGCA\cite{MGCA} and MRM \cite{MRM}.  We only report detection results on ResNet since we use YOLOv3\cite{YOLOv3} as our backbone, which is based on the convolutional neural network. Despite being pre-trained on a relatively small dataset, our method demonstrates balanced and nearly the best performance across various tasks, while other methods exhibit some shortcomings. Numerical details are in the supplementary materials.}
    \label{multimodal}
\end{figure}

\begin{figure}[t!]
    \centering
    \includegraphics[scale=0.33]{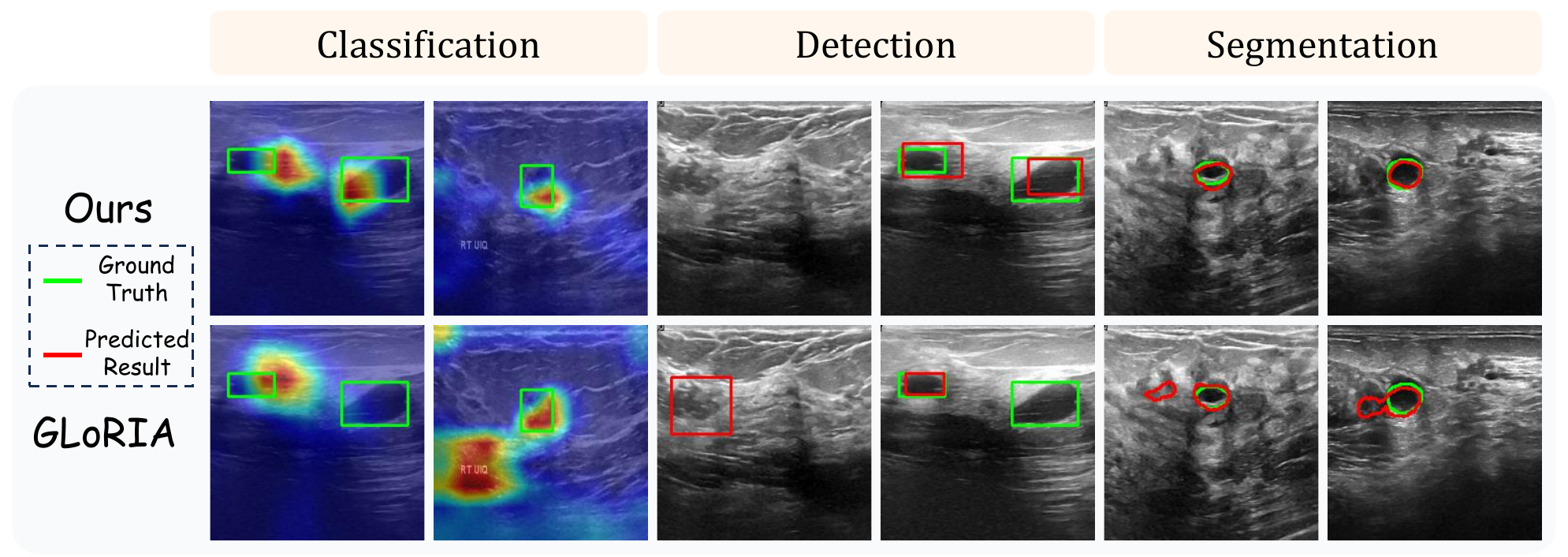}
    \caption{Examples of visual recognition tasks. Our method achieves superior performance in both classification, detection, and segmentation compared to GLoRIA.
    }
    \label{vision visualization}
\end{figure}

\subsubsection{Visual recognition.}


In this section, we compare our model's performance in three visual recognition tasks with that of other SOTA multimodal pre-training methods in Figure \ref{multimodal}. We observe that most of the previous SOTA methods struggle to achieve balanced performance across various downstream tasks when pre-trained on our dataset which is relatively smaller than their pre-training dataset. For instance, MRM \cite{MRM} achieves excellent segmentation performance but performs poorly in classification. Our approach achieves competitive performance across all tasks. We believe that this is because VQA enhances the utilization efficiency of images. Additionally, our multi-granular VQA enables the model to learn to extract multi-granular features, thus benefiting various downstream tasks. Figure \ref{vision visualization} visualizes some of the examples. Compared to GLoRIA \cite{GLoRIA}, our method achieves more precise recognition and less misidentification.

\begin{table}[t!]
\centering
\caption{Performance of report generation. Our method shows the best in most cases compared to other SOTA methods. B1 to B4 represent BLEU-1 to BLEU-4, while MR and RL represent METEOR and ROUGE-L, respectively.}
\label{report generation}
\begin{tabular}{>{\centering\arraybackslash}m{1.1cm}|p{2.2cm}>{\centering\arraybackslash}m{1.3cm}>{\centering\arraybackslash}m{1.3cm}>{\centering\arraybackslash}m{1.3cm}>{\centering\arraybackslash}m{1.3cm}>{\centering\arraybackslash}m{1.4cm}>{\centering\arraybackslash}m{1.5cm}}
\hline
\multicolumn{1}{c|}{Dataset} & Method        & B1$\uparrow$          & B2$\uparrow$          & B3$\uparrow$          & B4$\uparrow$          & MR$\uparrow$          & RL$\uparrow$         \\ \hline
\multirow{5}{*}{Breast}  
                         & TriNet\cite{TriNet}  & 0.693 & 0.594 & 0.533 & 0.478 & 0.439 & 0.742 \\
                         & R2Gen\cite{R2Gen}   & 0.663 & 0.611 & 0.572 & 0.541 & 0.411 & 0.685 \\
                         & TF\cite{transformer}      & 0.699 & 0.653 & 0.619 & 0.590 & 0.437 & 0.757 \\ 
                         & R2GenRL\cite{R2GenRL}      & 0.616 & 0.528 & 0.464 & 0.414 & 0.470 & 0.599 \\
                         & DeltaNet\cite{DeltaNet}      & 0.716 & 0.661 & 0.628 & 0.598 & \textbf{0.517} & 0.758 \\ \cline{2-8} 
                             & \textbf{Ours} & \textbf{0.730} & \textbf{0.695} & \textbf{0.665} & \textbf{0.639} & 0.442 & \textbf{0.774} \\ \hline
\multirow{5}{*}{Thyroid} 
                         & TriNet\cite{TriNet}  & 0.645 & 0.510 & 0.421 & 0.345 & 0.409 & 0.678 \\
                         & R2Gen\cite{R2Gen}   & 0.578 & 0.532 & 0.492 & 0.457 & 0.369 & 0.664 \\
                         & TF\cite{transformer}      & 0.709 & 0.642 & 0.585 & 0.538 & 0.425 & 0.701 \\
                         & R2GenRL\cite{R2GenRL}      & 0.672 & 0.595 & 0.531 & 0.479 & \textbf{0.500} & 0.651 \\
                         & DeltaNet\cite{DeltaNet}& 0.610 & 0.559 & 0.515 & 0.479 & 0.443 & 0.686\\ \cline{2-8} 
                             & \textbf{Ours} & \textbf{0.755}  & \textbf{0.712}  & \textbf{0.674}  & \textbf{0.640}  & 0.444  & \textbf{0.761}  \\ \hline
\end{tabular}
\end{table}

\subsubsection{Report Generation.}
We compare our method with other SOTA report generation methods in Table \ref{report generation}. Our method achieves the best performance in most cases. Specifically, our method outperforms the suboptimal model (DeltaNet \cite{DeltaNet} and TF \cite{transformer}) by 6.856\% on the breast dataset and by 18.95\% on the thyroid dataset in BLEU-4. We assume that the benefit arises from the design of medium and fine-grained VQA which enhances the precision of nodule recognition, and the QFT module contributes to generating more accurate reports. Additionally, we observe that our method does not achieve the highest METEOR score, which is more related to the correctness of token order. However, we believe that the order for describing different anatomical regions in the medical report is less crucial. Our method pays more attention to pathological features rather than the order of each region, resulting in higher BLEU and ROUGE-L scores.



\begin{table}[t!]
\caption{Results of ablation studies. P and R denote precision and recall for nodule recognition in reports. The introduction of QFT improves the quality of report generation, while the addition of VQA reduces the misidentification rate of nodules. }
\centering
\begin{tabular}{>{\centering\arraybackslash}m{1cm}>{\centering\arraybackslash}m{0.6cm}>{\centering\arraybackslash}m{0.5cm}>{\centering\arraybackslash}m{0.7cm}|>{\centering\arraybackslash}m{1.1cm}>{\centering\arraybackslash}m{1.1cm}>{\centering\arraybackslash}m{1.1cm}>{\centering\arraybackslash}m{1.1cm}>{\centering\arraybackslash}m{1.3cm}>{\centering\arraybackslash}m{1.2cm}>{\centering\arraybackslash}m{0.6cm}>{\centering\arraybackslash}m{0.6cm}}
\hline
\begin{tabular}[c]{@{}c@{}}Multi\\Images\end{tabular} &QCL &CL &VQA &
  \begin{tabular}[c]{@{}c@{}}B1$\uparrow$\end{tabular} &
  \begin{tabular}[c]{@{}c@{}}B2$\uparrow$\end{tabular} &
  \begin{tabular}[c]{@{}c@{}}B3$\uparrow$\end{tabular} &
  \begin{tabular}[c]{@{}c@{}}B4$\uparrow$\end{tabular} &MR$\uparrow$ &
  \begin{tabular}[c]{@{}c@{}}RL$\uparrow$\end{tabular} &
  \begin{tabular}[c]{@{}c@{}}P$\uparrow$\\ (\%)\end{tabular} &
  \begin{tabular}[c]{@{}c@{}}R$\uparrow$\\ (\%)\end{tabular} \\ \hline
  &   &   &   & 0.652          & 0.619          & 0.591          & 0.568          & 0.396          & 0.717          & -             & -            \\
\checkmark &   &   &   & 0.703          & 0.662          & 0.627          & 0.598          & 0.416          & 0.743          & -             & -   \\
\checkmark & \checkmark &   &   & 0.678          & 0.641          & 0.609          & 0.582          & 0.407          & 0.722          & -             & -            \\
\checkmark & \checkmark & \checkmark &   & 0.728          & 0.692          & 0.660          & 0.632          & 0.437          & 0.767          & 85.0            & 94.0           \\
\checkmark & \checkmark & \checkmark & \checkmark & \textbf{0.730} & \textbf{0.695} & \textbf{0.665} & \textbf{0.639} & \textbf{0.442} & \textbf{0.774} & \textbf{87.5} & \textbf{100} \\ \hline
\end{tabular}
\label{RG ablation}
\end{table}

\subsubsection{Ablation Study.}
In this section, we show the ablation study of our framework on the breast report generation task in Table \ref{RG ablation}. Firstly, although only using the most relevant image is a common approach in previous research \cite{CNN-LSTM,R2Gen,TriNet}, we observe that using multiple images as input improves the quality of the generated reports (0.717 vs 0.743). As mentioned in Sec. \ref{QFT section}, adding QCL results in a trivial solution (0.722), while the addition of CL leads to a notable increase (0.767), demonstrating that the constraint of CL can effectively reduce the loss of image information during QCL. The addition of VQA further improves the performance (0.767 vs 0.774). Besides, we conduct further analysis to evaluate the impact of VQA: we randomly sample 100 reports and calculate the precision and recall for nodule recognition. Specifically, we regard it as a binary classification task, where reports mentioning the presence of nodules are labeled as 1, and otherwise as 0. We observe a significant improvement (85.0\% vs 87.5\% on precision, 94.0\% vs 100\% on recall). This is crucial in the medical field as it can reduce misdiagnosis.

\section{Conclusion}
In this paper, we take the lead in exploring the potential of VQA in pre-training. We design different levels VQA targeting vital pathological features according to the description in the medical report without any extra annotations from clinicians. We also propose a pre-training framework with QFT, a module used to narrow the vision-language gap with a contrastive learning strategy. We demonstrate the effectiveness of our approach in report generation and three visual recognition tasks. Experimental results indicate that VQA guides the model focusing on desired pathological features, demonstrating the potential of VQA in pre-training. This work is an initial exploration. We will further investigate how to design reasonable questions and how to efficiently utilize VQA in pre-training. 

\subsubsection{Acknowledgement.} This work was supported by the Key Fundamental Research Program of Shenzhen (No. JCYJ20220818101408019).

%
%
%
\bibliographystyle{splncs04}
\bibliography{refs.bib}

\newpage
\title{Supplementary Material for Design as Desired: Utilizing Visual Question Answering for Multimodal Pre-training}

\author{Tongkun Su\inst{1,2,*} \and
Jun Li\inst{3,}\thanks{The first two authors contributed equally to this work.} \and
Xi Zhang\inst{1,2} \and Haibo Jin\inst{5} \and Hao Chen\inst{5} \and \\ 
Qiong Wang\inst{1,6} \and Faqin Lv\inst{4} \and Baoliang Zhao\inst{1(}\Envelope\inst{)} \and Yin Hu\inst{1(}\Envelope\inst{)}} 
\institute{Shenzhen Institute of Advanced Technology, Chinese Academy of Science
\and
University of Chinese Academy of Science
\and
Technical University of Munich, Munich Center for Machine Learning 
\and
Southern Medical University
\and
The Hong Kong University of Science and Technology
\and
The Chinese University of Hong Kong
}
\maketitle

\begin{figure}
    \centering
    \includegraphics[width=\textwidth]{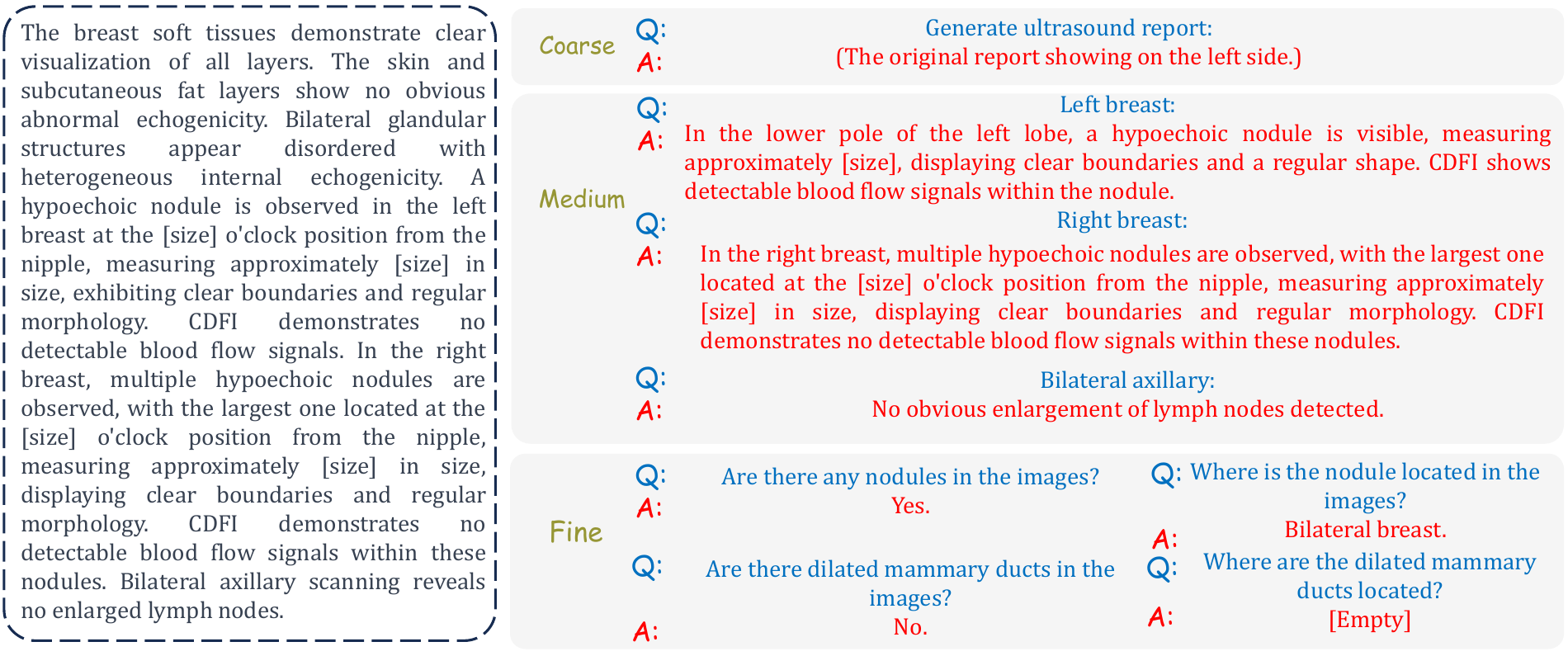}
    \caption{Example of the breast VQA design. [Empty] indicates that this question will not be treated as pre-training data.}
    \label{VQA example}
\end{figure}



\begin{table}[]
\centering
\caption{Configuration of pre-training.}
\begin{tabular}{c|cc}
\hline
Configuration         & Multimodal pretraining & Report generation     \\ \hline
Optimizer             & AdamW                  & AdamW                 \\
Learning rate         & 2e-5                   & 2e-5                  \\
Weight decay          & 0.05                   & 0.05                  \\
Learning rate scheduler &
  \begin{tabular}[c]{@{}c@{}}Linear warmup+\\ cosine annealing\end{tabular} &
  \begin{tabular}[c]{@{}c@{}}Linear warmup + \\ cosine annealing\end{tabular} \\
Initial learning rate & 1e-8                   & 1e-8                  \\
Warmup periods        & 40\% of training time  & 40\% of training time \\
Early stop            & 5                      & 5                     \\
Batch size(B)         & 25                     & 25                    \\
Buffer size(N)        & 100                    & 100                   \\
Query(m)              & 32                     & 32                    \\
Epoch &
  \begin{tabular}[c]{@{}c@{}}30 for ViT and \\ 50 for ResNet\end{tabular} &
  50 \\ \hline
\end{tabular}
\end{table}

\begin{table}[]
\caption{Finetuning configuration of different downstream vision tasks.}
\centering
\begin{tabular}{c|ccc}
\hline
Configuration & Classification & Detection & Segmentation \\ \hline
Optimizer     & AdamW          & AdamW     & AdamW        \\
Early stop    & 10             & 10        & 10           \\
Epoch         & 50             & 50        & 50           \\
Method        & Linear probe   & YOLOv3    & SETR/UNet    \\
Learning rate & 5e-4           & 5e-4      & 2e-4         \\
Weight decay  & 1e-6           & 1e-6      & 0.05         \\
Batch size    & 48             & 16        & 8            \\ \hline
\end{tabular}
\end{table}

\begin{table}[]
\caption{Quantitative result of multimodal pre-training. The results have been presented in the main text. \protect\say{-} indicates that it is not suitable for this situation.}
\centering
\begin{tabular}{lcccccc}
\hline
\multirow{2}{*}{Method} & \multicolumn{2}{c}{Classification(AUC\%)} & \multicolumn{2}{c}{Detection(AP\%)} & \multicolumn{2}{c}{Segmentation(DICE\%)} \\
                   & BUSI           & AUITD          & BUSI           & DDTI           & BUSI           & DDTI          \\ \hline
Random(ViT)        & 56.4          & 81.3          & -              & -              & 38.1          & 58.1         \\
Random(Res)        & 61.5          & 81.3          & 51.5          & 13.9          & 58.0          & 64.7         \\
ImageNet(ViT)      & 84.5          & 82.5          & -              & -              & 63.9          & 61.5         \\
ImageNet(Res)      & 82.9          & 82.2          & \textbf{66.7} & 50.0          & 49.0          & 61.1         \\
GloRIA(Res)        & 85.5          & 80.2          & 54.9          & 21.1          & 63.7          & 63.8         \\
MGCA(ViT)          & 82.9          & 80.2          & -              & -              & 61.2          & 68.7         \\
MGCA(Res)          & 82.9          & 82.2          & 55.5          & 10.5          & 59.2          & 68.8         \\
MRM(ViT)           & 69.2          & 81.3          & -              & -              & 61.1          & \textbf{73.1} \\ \hline
\textbf{Ours(ViT)} & \textbf{88.9} & \textbf{83.3} & -              & -              & 63.5          & 70.2         \\
\textbf{Ours(Res)} & 84.6          & 82.2          & 62.1          & \textbf{57.9} & \textbf{65.6} & 70.4         \\ \hline
\end{tabular}
\label{MMP}
\end{table}

\end{document}